\documentclass[10pt, a4paper]{article}
\usepackage{lrec}
\usepackage{graphicx}
\usepackage{tabularx}
\usepackage{soul}

\usepackage{epstopdf}
\usepackage[latin1]{inputenc}

\usepackage{hyperref}
\usepackage{xstring}

\usepackage{times}
\usepackage{latexsym}
\usepackage{booktabs}
\usepackage{amssymb}
\usepackage{bm}
\usepackage{todonotes}
\usepackage{url}

\usepackage{enumitem}

\newcommand{\secref}[1]{\StrSubstitute{\getrefnumber{#1}}{.}{ }}

\newcommand{\ra}[1]{\renewcommand{\arraystretch}{#1}}

\title{A Corpus of Controlled Opinionated and Knowledgeable Movie\\ Discussions for Training Neural Conversation Models\\ \vspace*{.5\baselineskip}}

\name{Fabian Galetzka\textsuperscript{1,2}, Chukwuemeka U. Eneh\textsuperscript{2}, David Schlangen\textsuperscript{1}}

\address{ \textsuperscript{1}Computational Linguistics, University of Potsdam, Germany\\
          \textsuperscript{2}Volkswagen AG\\
         fabian.galetzka@volkswagen.de}

\abstract{
    Fully data driven Chatbots for non-goal oriented dialogues are known to suffer from inconsistent behaviour across their turns, stemming from a general difficulty in controlling parameters like their assumed background personality and knowledge of facts.
    One reason for this is the relative lack of labeled data from which personality consistency and fact usage could be learned together with dialogue behaviour. To address this, we introduce a new labeled dialogue dataset in the domain of movie discussions, where every dialogue is based on pre-specified facts and opinions. We thoroughly validate the collected dialogue for adherence of the participants to their given fact and opinion profile, and find that the general quality in this respect is high. This process also gives us an additional layer of annotation that is potentially useful for training models. We introduce as a baseline an end-to-end trained self-attention decoder model trained on this data and show that it is able to generate opinionated responses that are judged to be natural and knowledgeable and show attentiveness. \\ \newline  }

\begin{document}

\maketitleabstract

\section{Introduction}
\label{main_introduction}
Where dialogue modelling used to be mostly rule-based with the dialogue being driven by pre-specified knowledge representations (e.g., \cite{Bobrow1977}, \cite{traumlarsson:ISbook}, \cite{stedeschlang:catabot}), recent years have seen efforts of basing this task on models directly learned from data. A particular strand of this research has modelled the task of producing a dialogue contribution in analogy to the translation task as one of going from one sequence (the user utterance) to another sequence (the system utterance). 

\begin{figure}[!h]
\begin{center}
\includegraphics[scale=0.6]{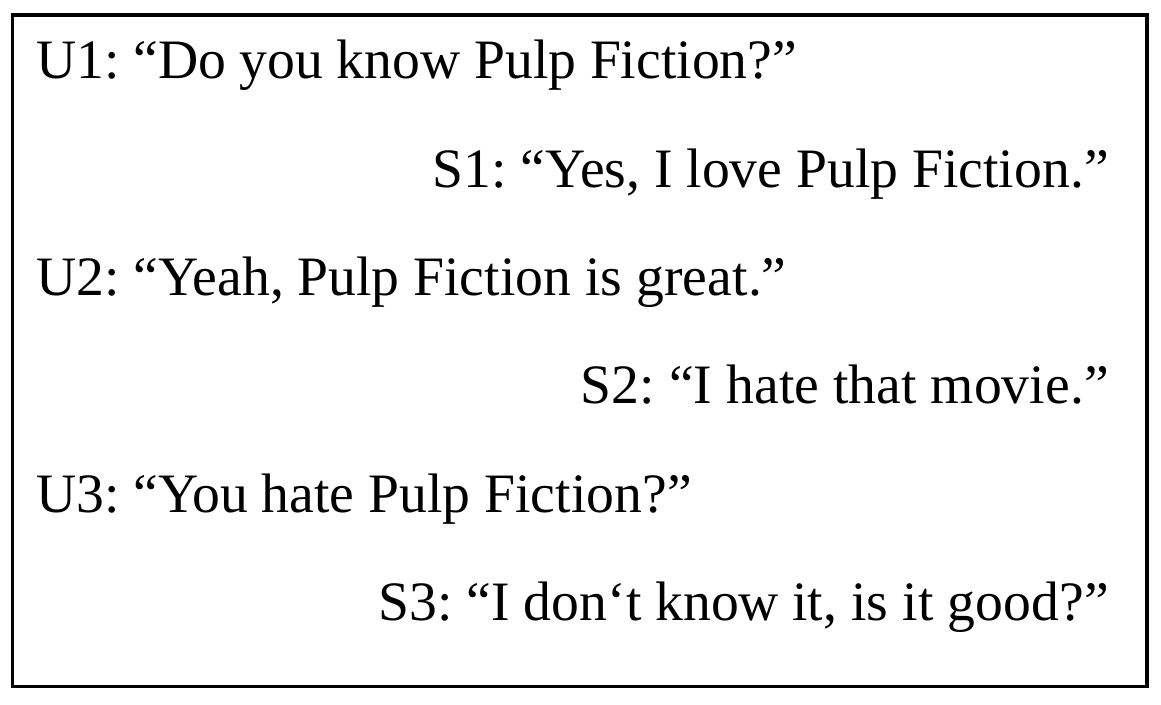} 
\caption{A constructed example of a dialogue that is \textit{locally} coherent, but \textit{globally} incoherent along the dimensions \emph{knowledge} (S3 to S1, S2) and \emph{opinion} (S2 to S1).}
\label{fig.bad_dialogues}
\end{center}
\end{figure}

The first such models solely based on data driven end-to-end approaches \cite{li2016deep,serban2017hierarchical} tended to generate universal and inconsistent utterances regarding content and personality. We illustrate this problem with the example in Figure~\ref{fig.bad_dialogues}, distinguishing the two consistency dimensions \emph{knowledge} (a speaker should not ``forget'' previously known facts) and \emph{opinion} (a speaker should not change their opinion, at least not without any overt trigger in the conversation). In this example, each system response is locally coherent (a good reply to its immediate precursor), but globally inconsistent.

While this particular example is constructed, it is not very far from what these early models would have been liable to produce. One reason for this is that these models were optimised only for local coherence, and trained from datasets such as TwitterCorpus~\cite{ritter2010unsupervised} and OpenSubtitles corpus~\cite{tiedemann2012parallel}. These datasets contain dialogues from many people, without any information about the speakers and their opinions or knowledge state.

To tackle issues like these, several augmented dialogue datasets have been introduced in recent years. \newcite{zhou2018dataset} created a dataset with conversations based on Wikipedia articles about popular movies. Another more general dataset  \cite{dinan2018wizard} explicitly tasked one person in each conversation to link the used knowledge to each written utterance. Models trained on these augmented datasets produced to more engaging and more natural dialogues, as shown in that paper. As opposed to additional general knowledge, the \textsc{persona-chat} dialogue corpus~\cite{zhang2018personalizing} is based on personality profiles. Crowd workers were matched together in a role-playing chat and asked to get to know each other, considering profile information which was individually provided for every participant. Different types of neural networks were trained on that dataset, which were shown to also produce more engaging and consistent dialogues compared to models trained on other datasets.

\begin{figure*}[!t]
\begin{center}
\includegraphics[scale=0.8]{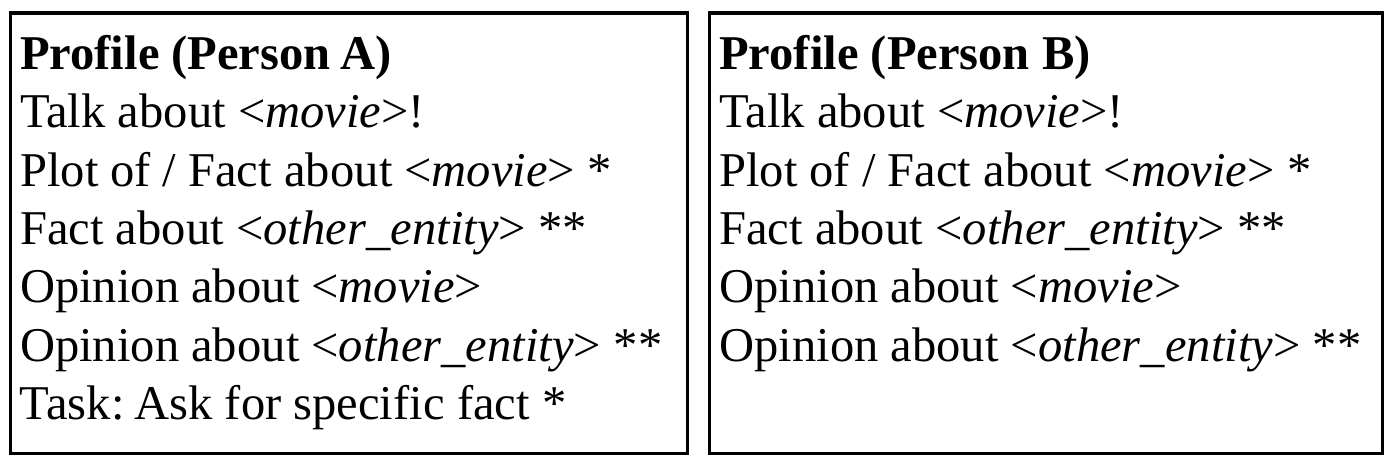} 
\caption{An abstract profile pair $P_\mathrm{A}, P_\mathrm{B}$, each given to one chat partner. * denotes information that may not occur in every profile. ** denotes the possibility of multiple occurrences with different entities.}
\label{fig.mturk_abstract}
\end{center}
\end{figure*}

We contribute to this research a corpus that combines these strands, as it consists of dialogues that were collected in a setting where we controlled both the knowledge available to the dialogue participants, as well as their stance towards entities to be mentioned in it. We present this corpus in the next section, and then show that from it models can be induced that better exhibit global consistency, along these dimensions.

\section{The \textsc{KOMODIS}-Dataset: Collection}
\label{main_data}

We introduce a new augmented dialogue dataset (\textbf{K}nowledgable and \textbf{O}pinionated \textbf{MO}vie \textbf{DIS}cussions) that is crowd-sourced and collected with Amazon Mechanical Turk (AMT). Every dialogue is constrained by a unique set of facts as well as a suitable set of opinions about the entities in the facts. The dataset is set in the domain of movies, which we selected because it is a popular topic likely to generate engaging conversations. 
The creation of the dataset is described in the present section, its validation relative to the aims of controlling opinion background and knowledgeability is described in detail in Section~\ref{data_validation}
The dataset is publicly available in our online repository\footnote{\url{https://github.com/fabiangal/komodis-dataset}}.

\begin{figure*}[!t]
\begin{center}
\includegraphics[scale=0.6]{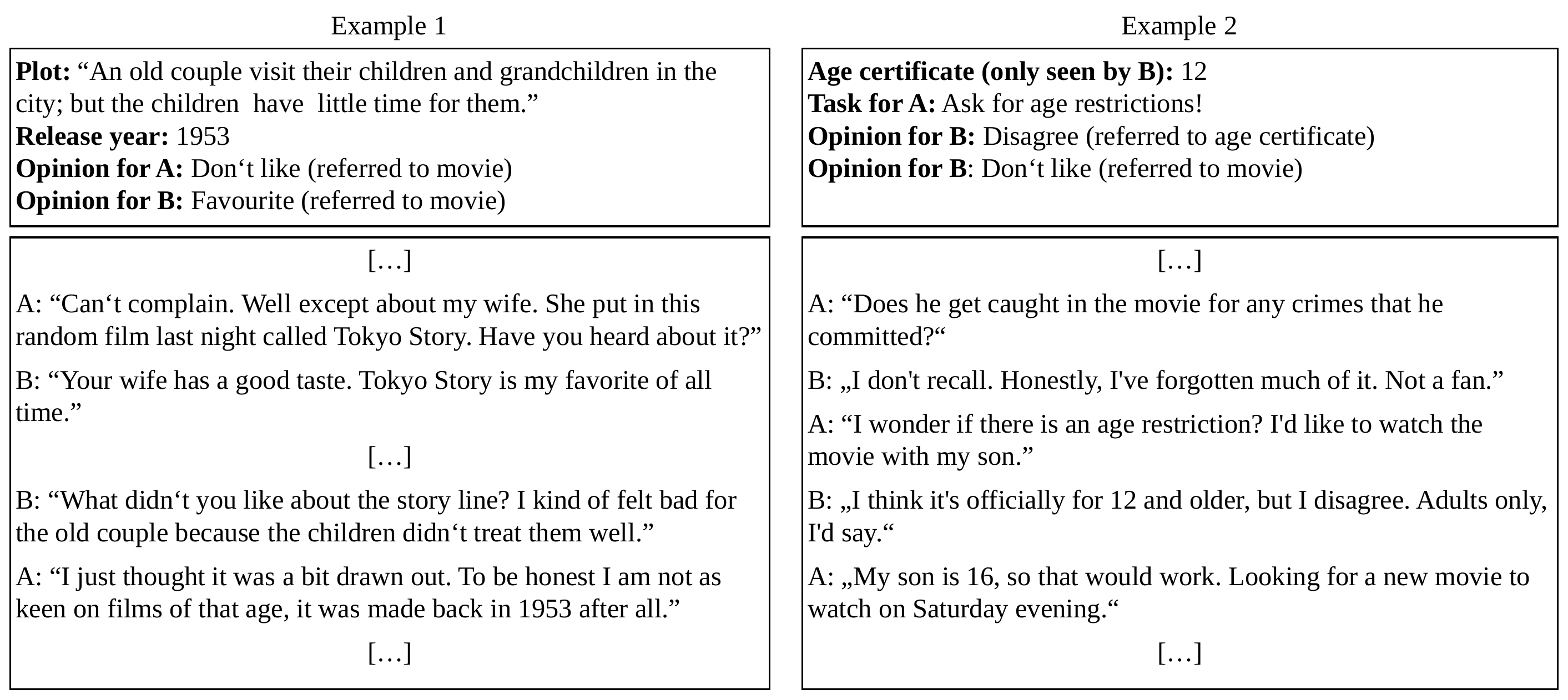} 
\caption{Parts of two dialogues from our corpus with the given relevant facts and attitudes. One can see that both participants combined the given information within the conversation with success. And even though the topic is constrained, the dialogues have open chit-chat characteristics.}
\label{fig.example_dialogues_dataset}
\end{center}
\end{figure*}

Inspired by \cite{zhang2018personalizing} our dialogues are collected by providing additional information to the participants. Unlike in that work, however, we do not indicate a textually-described personality, but rather provide facts about entities (knowledge) and opinions about them. For each dialogue we created a unique set of two \emph{profiles} (formalised here as feature structures).
In all cases both crowd-worker had to talk about the same movie, with different combinations of feature structures $P_\mathrm{A}, P_\mathrm{B}$. An abstract example is shown in figure \ref{fig.mturk_abstract}. The facts are explained in more detail in \ref{data_creation_facts}, as well as the opinions in \ref{data_creation_opinions} The different combinations of feature structures are explained in \ref{data_creation_unification} A concrete example is shown in Figure~\ref{fig.example_dialogues_dataset}.

\subsection{Facts}
\label{data_creation_facts}
The facts are a combination of three different types of information, all extracted from the publicly available movie database IMDb\footnote{\url{https://www.imdb.com/}}: 

(1) Open-domain sentences, so called trivia about movies and actors. For 
example: `\textit{The screenplay says that Zed and Maynard are brothers.}', or: `\textit{Quentin Tarantino was quoted as saying that Butch is responsible for keying Vincent's car.}'. These trivia information is itself crowd-sourced in IMDb, but comes with a crowd-sourced rating. We only use such trivia marked as \emph{interesting} in the IMDb. We also used the overall length of the trivia, with shorter trivias preferred over longer ones, to ensure a compact set of facts in the end.

(2) A short plot of every movie. For example from Pulp Fiction: 
`\textit{The lives of two mob hitmen, a boxer, a gangster's 
    wife, and a pair of diner bandits intertwine in four tales of violence and 
    redemption.}'

(3) Facts like release date or budget of a movie. While the trivia have the form of open-domain sentences, these facts are given as knowledge triples in the database. We created multiple sentence patterns per type of fact to convert them into sentences as well.  

Given a specific movie, we took 2--4 facts to generate a set. The facts were chosen randomly with a few constraints to ensure a fluent dialogue. For example, if a randomly selected trivia about a movie mentioned an actor, the next fact could be about that actor and so on:

(1) Sometimes one participant is asked to pretend not to know a certain movie, in which case they do not get any information about it. Instead we provide at least one question.

(2) If one participant gets the task to ask a specific question, we provide the correct answer to the other participant.

(3) We prioritized trivias that include entities of actors from the given movie. If that is the case, we provided additional information about this actor.

(4) We randomly added additional facts like budget or genre, but not every set of facts has one of these information.

(5) Every trivia is only used once in the whole dataset.

\subsection{Opinions}
\label{data_creation_opinions}
We augmented the facts by a set of suitable opinions. For example, if a trivia is about an actor, we provided an opinion about that actor. We used a discrete set of opinions ranging from \textit{really don't like} to \textit{favorite} as well as \textit{don't know}. The attitudes were converted into sentences, too. Their strength was generated randomly and all possible combinations are available.

\subsection{Relations between Speaker Profiles}
\label{data_creation_unification}

To induce interesting dialogues, we varied the relations between the profiles. In the first type of relation, both have the same profile (knowledge of facts and opinions about them):
\begin{equation}
\label{unification_equal}
    P_\mathrm{A} = P_\mathrm{B}
\end{equation}

We also create profile sets where the individual profiles are complimentary, but not conflicting (e.g., A knows something that B doesn't; formally, the feature structures representing the profiles can be unified):
\begin{equation}
\label{unification_compatible}
    P_\mathrm{A} \sqcup P_\mathrm{B}
\end{equation}

Finally, we also created sets with incompatibilities (only along opinions, however, since we did not want them to get into factual discussions):
\begin{equation}
\label{unification_conflict}
    P_\mathrm{A} \sqcup P_\mathrm{B} = \perp
\end{equation}

\subsection{Collection Procedure}
\label{data_amt}
The dataset was collected via Amazon Mechanical Turk and with the help of 
\textit{slurk}, a chat server developed for data collection tasks that require a matching of multiple participants 
\cite{schlangen2018slurk}. Two crowd-worker were paired and tasked to chat using the 
provided information (different for each of them). The crowd-worker were 
tasked to use both the facts and opinions to generate 
sensible utterances within the conversation.

AMT provides two types of payments. A basic one, which is fixed and bound to the task and a flexible bonus payment, that can be paid manually afterwards. Matching two participants requires at least one of them to wait for a partner. We used that process of waiting for the small basic payment. Then, after a successful match, we paid most of the fee for up to three minutes of chatting as bonus payment. If crowd-worker waits for three minutes, they can finish the task without chatting; this happened in less then 5\% of the cases though.  

In our first iterations we figured out that the crowd-worker tended to simply copy the trivia and rush through the facts. Another problem with written chat is that cross talk can occur (where both participants are typing at the same time and messages get out of order). We found that by enforcing who started the conversation, by giving one randomly selected participant the first turn, we could reduce this, without having to enforce strict turn taking throughout the interaction. This increased the data quality considerably. Also, the quality of the dialogues increased with the amount of money we paid. A bonus payment for 'well written utterances' also helped. We paid up to $1.60\$$ per dialogue. Additionally we limited the number of tasks one crowd-worker can do per day by 5.

After the chat we asked the participants to rate their partner in terms of language quality, naturalness and attentiveness.\footnote{Actual statements the participants had to rate: ``My partner is a native speaker of English'',``This felt like a natural chat about movies'',``My partner chatted like an attentive person would''} We speculated that this information might be useful to detect bad quality dialogues, and could also serve as a baseline for human evaluation of trained models.

\section{Dataset Overview and Validation}
In the following section we present a quantitative overview of our dataset, as well as a detailed validation of the data.

\subsection{Dataset Statistics}
\label{data_statistics}
We initially collected $8,000$ interactions. From these, we had to filter out 1,032 ($12.9\%$) because either one participant did not respond in the conversation or one participant rated his partner's quality negatively. In a second iteration we collected another batch, bringing the total up to $7,519$ dialogues. In these, there is an average number of $13.8$ speaker turns ($103,500$ in total). We have split our dataset into a train, validation and test set with $80\%$, $10\%$ and $10\%$ dialogues respectively, in such a way to no movie is in more than one split. We give some descriptive statistics about the dataset in table \ref{t_dataset}.

\begin{table}[t]
\ra{1.2}
    \begin{center}
        \begin{tabular}{lr}
            \toprule[1.5pt]
            \bf Parameter & \bf Value \\ \hline
            dialogues & $7,519$\\
            utterances & $103,500$ \\
            tokens & $1,487,284$ \\
            average utterances per dialogue & $13.8$ \\  
            average tokens per utterance & $14.4$ \\ \hline
            vocabulary size &  $27,658$\\
            vocabulary size (99\% of tokens) & $13,727$ \\ \hline
            different movies & $500$  \\
            used (unique) trivia  & $13,818$  \\ \hline
            participants from AMT  & $569$  \\
            \bottomrule[1.5pt]
        \end{tabular}
    \end{center}
    \caption{\label{t_dataset} Quantitative representation of our dataset. Trivias and movies are approximately evenly distributed over all dialogues. The vocabulary size was computed separately by counting all different tokens from the original dialogues. However, for training we used byte pair encoding.}
\end{table}

\subsection{Dataset Validation}
\label{data_validation}

After collecting the dialogues we post-processed and validated the dataset. As it is not possible to supervise the crowd-worker automatically while chatting, we have to be sure that a) they really talked about the profile entities and  b) adhere to the opinions specified there. 

\subsubsection{Named Entity Resolution}
As a first step we extracted all named entities from each dialogue. Even though with the existence of powerful natural language processing tools like Spacy \cite{spacy2} and CoreNLP \cite{corenlp}, which can detect mentions of names, organizations or countries with high precision (named entitiy recognition, NER), detecting movie titles still remains a challenging problem \cite{ashwini2014targetable}, especially with grammatical errors and spelling mistakes. However, for each dialogue, we knew which movie they were (supposed to be) chatting about, which reduces the complexity of named entity recognition in our domain. We used three different metrics to find an entity: First, exact string match on the lowercased strings, which has high precision but very low recall. Second, we compared every n-gram in an utterance and the movie title with the cosine similarity from Spacy. We used a threshold of $0.9$ and
\begin{equation}
\min{(t_{\mathrm{movie}}; 3)} \leq n \leq t_{\mathrm{movie}}
\end{equation}
for the n-grams, with $t_{\mathrm{movie}}$ as the number of tokens of a movie title. And third, a combination of the Jaccard distance \cite{niwattanakul2013using} with threshold $\leq 3$ and Levenshtein distance \cite{levenshtein1966binary} with threshold $\leq 0.3$ for the same n-grams. For mentioned persons we used the pretrained named entity recognition from Spacy in addition to the aforementioned metrics.

To evaluate our automatic algorithm, we randomly chose 50 dialogues and asked an assistant who was not otherwise involved in the project to manually annotate these dialogues. On this, admittedly small, set our automatic method reached high NER precision and recall with $97.8\%$ and $91.9\%$ respectively. The lower recall is mostly caused by typing errors from the crowd workers, so that our algorithm could not detect some of the entities.

\subsubsection{Usage of Profile Entities}
To show that the crowd-worker really talked about the given profile entities, we computed the overall coverage of named entities. For every dialogue we compared the entities given to the worker in the profile and the detected named entities in the dialogue; counting each match. Averaging over the dialogues, we find that $93.1\%$ of the profile entities are indeed mentioned in a dialogue. (We did not calculate whether \textit{additional} entities may have been mentioned, as we did not want to restrict that from happening.)

\subsubsection{Adherence to the Opinion Profiles}
Another crucial property is the correct usage of the given opinions. Automatically validating this was not trivial, as it requires co-reference resolution and sentiment analysis for our specific data. We assumed that the effort would be worthwhile, though, as a detailed sentiment analysis would augment the dataset with additional fine-graded information potentially useful for training models with the data (see next section).

\begin{table}[t]
\ra{1.2}
    \begin{center}
        \begin{tabular}{lr}
            \toprule[1.5pt]
            \bf named entity resolution \\
            precision & $0.977$ \\
            recall & $0.920$ \\
            f1-score & $0.947$ \\ 
            \bottomrule[1.5pt]
        \end{tabular}
    \end{center}
    \caption{\label{t_ner} Evaluation of our named entity resolution.}
\end{table}

To detect an opinion about a named entity, we first had to resolve indirect references (e.g. ``\textit{I like it!}'' may need to be converted to ``\textit{I like Pulp Fiction!}''). We used the coreference resolution annotator from CoreNLP to replace the references with their entity names. First we substituted the original movie titles, person names, countries and genres (as recognised in the NER step) with placeholder words like ``\textit{Pulp Fiction}'' or ``\textit{Peter Pan}'' which we confirmed to be recognised by CoreNLP, as it turned out that unusual names or long movie titles are challenging for CoreNLP, especially with typos or lowercased. For our specific case we noticed some problems with the CoreNLP heuristics, presumably because our data is different from its training data. Therefore we manually filtered co-reference chains with first or second person pronouns, as CoreNLP had problems with resolving them correctly and in our case only third person entities are relevant. 

\begin{figure}[h!]
\begin{center}
\includegraphics[scale=0.58]{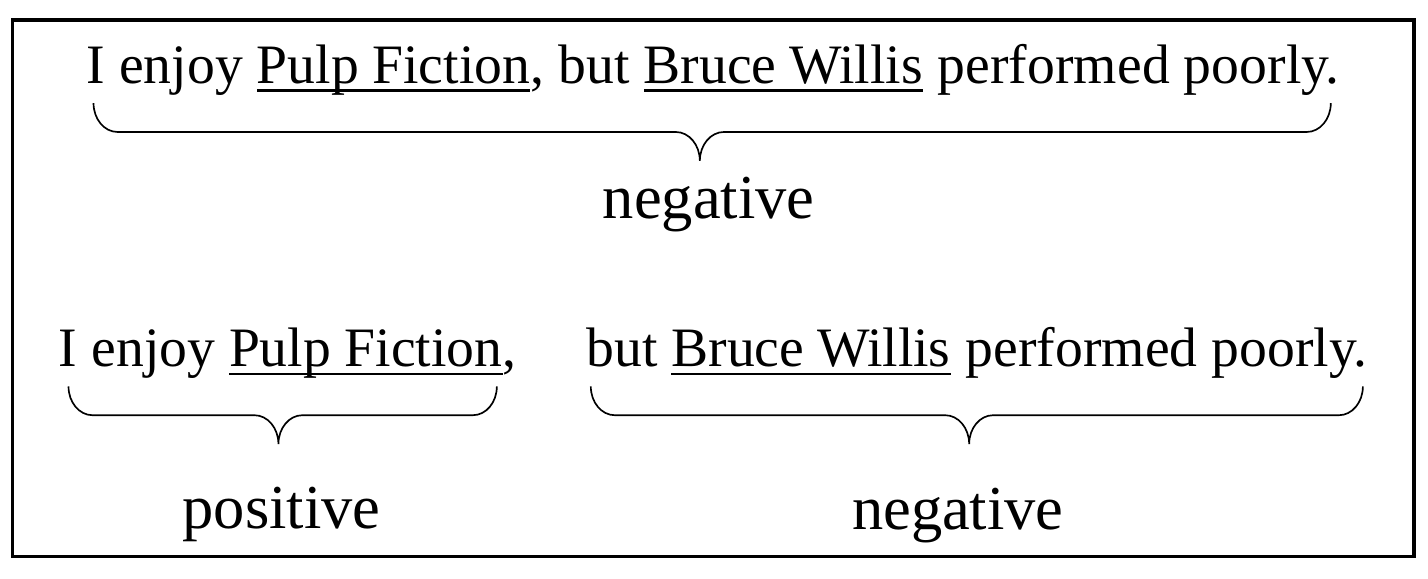} 
\caption{Estimated sentiments from CoreNLP. Named entities are underlined. In the first row Pulp Fiction is declared as negative by mistake.}
\label{fig.sentiment_sentence_splitting}
\end{center}
\end{figure}

To detect the named entity related sentiments, the smallest unit around an entity that can carry a sentiment needs to be identified. An example is given in figure \ref{fig.sentiment_sentence_splitting}. Therefore we used the annotated parse trees from CoreNLP and determined the smallest subordinate clauses within each sentence and all noun phrases with a recursive tree search. In a second step sentence fractions are merged until they contain up to two different nouns. We noticed problems with the annotated parse trees on sentences with grammatical errors, spelling mistakes or wrong punctuation marks, which led to low recall, as we had to ignore such sentences.
 
In a final step each subsentence was processed through the sentiment annotator from CoreNLP which provides a discrete probability distribution over five sentiments (\textit{VeryNegative, Negative, Neutral, Positive,  VeryPositive}). We compared these labels with the given opinions from the profiles.   

With that approach 53\% of all mentioned entities were labeled as neutral, in 80.1\% of the cases, the estimated sentiments conformed with the profile. For a meaningful evaluation of our dataset, the automated approach is not precise enough, so again we evaluated 50 randomly chosen dialogues manually.
The results of the manual evaluation are shown in table \ref{t_profileadherence}. For most the crowd-worker followed their instructions with a high accuracy of $97\%$.

\vspace*{\baselineskip}
To sum up, our analysis showed that the crowd-workers
\begin{enumerate}[label=(\roman*)]
\item produced relatively rich and long dialogues (on average, 14 turns), 
\item talked about the entities they were given as topics, and
\item took on the pre-specified opinions about them.
\end{enumerate}

\begin{table}[t]
\ra{1.2}
    \begin{center}
        \begin{tabular}{lrrrr}
            \toprule[1.5pt]
            \bf entity & \bf matches & \bf errors & \bf neutral & \bf accuracy \\ \hline
            movie & $122$ & $5$ & $36$ & $0.96$ \\
            person & $91$ & $1$ & $35$ & $0.99$ \\
            other & $57$ & $2$ & $28$ & $0.97$ \\ \hline
            sum & $270$ & $8$ & $99$ & $0.97$ \\ 
            \bottomrule[1.5pt]
        \end{tabular}
    \end{center}
    \caption{\label{t_profileadherence} Accuracy of crowd-worker adherence to their opinionated profile. Manually evaluated on 50 randomly chosen dialogues.}
\end{table}

\subsubsection{Detailed Sentiment Labels}
\label{data_detailedsentiments}
The validation of our dataset yielded a lot of useful information, which we use to augment our dataset with utterance-level labels regarding entities and sentiments. Later we show in section \secref{main_evaluation} that these labels can help to improve dialogue quality of our neural network models.

\section{Model}
\label{main_model}

\begin{figure}
    \begin{center}
        \includegraphics[width=\textwidth/3]{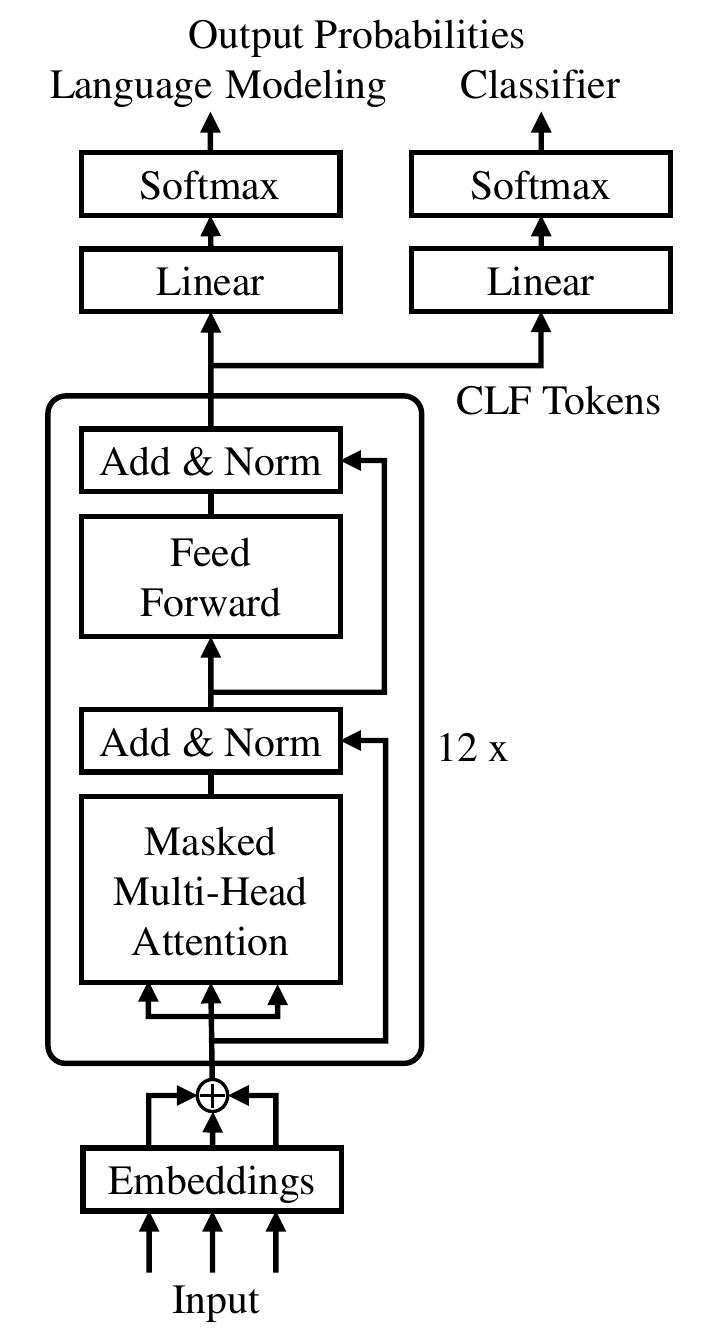}
        \caption{\label{fig.model_architecture} Transformer architecture  \protect\cite{radford2018improving} of our baseline model.}
    \end{center}
    \vspace*{-.7cm}
\end{figure}

\begin{figure*}
    \begin{center}
        \includegraphics[width=\textwidth]{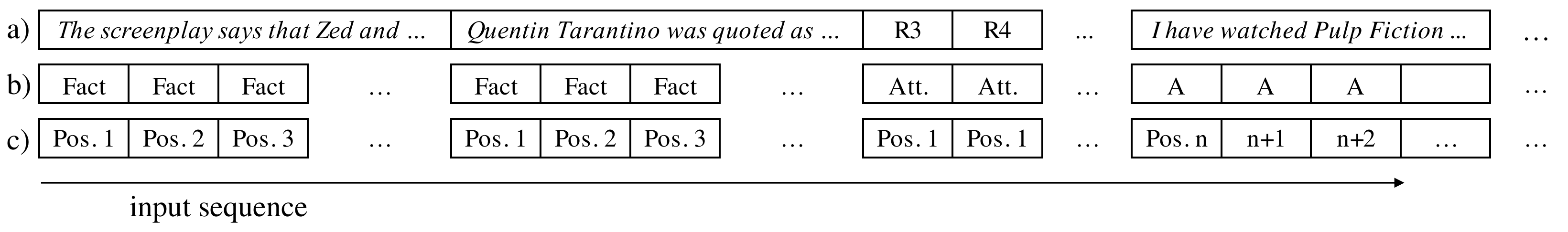}
        \caption{\label{fig.pos_emb} Input representation of our baseline model. The final embedding of each step in a sequence is the sum of all three embeddings: a) The byte paired encoding of the utterances. b) The content encoding that specifies the type of a token. c) The pre-trained positional encodings, shared across the facts.}
    \end{center}
\end{figure*}

To show the contribution of our dataset towards more controlled neural chat generation, we present a baseline model. The task for the model is to generate the next utterance, given a dialogue with some facts and opinions. It is a generative model trained on two objectives, language modeling with cross-entropy and next-utterance classification. It has the same architecture as the decoder part from \newcite{vaswani2017attention} with the multi-head attention over the encoder removed, as here we do not have an encoder. This architecture was introduced by \newcite{radford2018improving} and is commonly known as GPT. It has a stack of 12 identical layers and each layer consists of 2 sub-layers: a masked multi-head self-attention layer and a position-wise, fully connected feed-forward layer with residual connections \cite{he2016deep} around them. Like in the original implementation, we used 768 dimensional states and 3072 nodes in the feedforward layers and 12 heads per attention layer. It was used by \newcite{wolf2019transfertransfo} with great success in a similar task on the \textsc{persona} dataset and outperformed state-of-the-art approaches in the Conversational Intelligence Challenge 2.\footnote{\url{http://convai.io/}} Therefore we used this as our base. Our PyTorch implementation can be found in our repository, mentioned in Section~\ref{main_data}, as well. The model is diagrammed in Figure \ref{fig.model_architecture}.

\subsection{Pre-training} 
\label{model_pretraining}
We used generative pre-training \cite{radford2018improving} with a selfmade corpus inspired by \newcite{zhu2015aligning}. That corpus contains over 1 billion tokens from books across different genres.\footnote{Collection code will be included in the public repository.} The model weights $\Theta$ are trained  given the tokenized corpus $X=\{x_1, x_2, ..., x_n\}$ with minimizing the negative log likelihood: 
\begin{equation}\label{eq_nll}
    L_{\mathrm{lm}} = 
    \sum\limits_{i}{-\log{P(x_i|x_{i-1},...,x_{i-k};\Theta)}}
\end{equation}
which is a standard language modeling approach. The tokens are split into sequences with length $k=512$. This unsupervised task allows the model to learn long-term dependencies on coherent text, as well as the token and positional embeddings. 

\subsection{Training} 
\label{l_training}
Before fine-tuning we had to adapt our data so that it fits into the decoder architecture. Similar to \newcite{wolf2019transfertransfo}, we decided that for our baseline model, we concatenate facts, attitudes, dialogue history and the next utterance to one input sequence.

In contrast to the pre-training, our setup has a dialogue history, additional facts and attitudes instead of just concatenated sentences. Therefore we need additional input embeddings to represent our more structured data. We used a new set of embeddings which are added to the sum of word tokens and positional embeddings. We used them to differentiate whether a set of tokens (e.g.\ an utterance) belongs to a specific dialogue partner ('\textit{A}' or '\textit{B}'), a fact or an 
attitude. The latter are represented with additional embeddings, one for each discrete attitude state. The general concept is shown in Figure~\ref{fig.pos_emb}. Where '\textit{Fact}' and '\textit{Att}' are groups of tokens that differentiate between their targets (e.g.\ the movie or a specific person). To ensure invariance to the order of the facts and attitudes, the same positional embeddings are used across all additional input, which is also illustrated in Figure~\ref{fig.pos_emb}. Dialogue history, facts and attitudes are concatenated into sequences with a maximum length of 512 tokens. Furthermore we added a classification token at the end of the last utterance, which is ignored by the language modeling loss, but used as input for the classification loss. 

After the last hidden layer we multiplied all tokens that did not belong to the last utterance with zeroes to avoid the model learning to predict other tokens than the ones from the last utterance.

To improve generalization, we used delexicalisation for the named entities. That includes movie titles, actors, directors, writer, budget values, age certificates, genres, countries and release years. It is important to note that this step removes the possibility to talk about more than one movie at a time.

We have finetuned the model with a batchsize of $32$ for $240,000$ steps on our own dataset, which equals three epochs. After that, both the language modeling loss and the classification loss on our validation set stopped decreasing. A sequence has up to $512$ tokens with shorter sequences padded to the maximum sequence length. We used adam optimizier with an initial learning rate $lr = 6.25e-5$, $\beta_{1} = 0.9$, $\beta_{2} = 0.999$ and $\epsilon = 1e-8$. We reused most of the parameter from pre-training: General dropout after all layers with $p_{\mathrm{drop}}=0.1$, weight decay regularization \cite{loshchilov2017fixing} with $w=0.01$ and the new embeddings are initialized with simple weight initialization of $N(0,0.02)$.

\begin{table*}[t!]
    \ra{1.2}
    \begin{center}
        \begin{tabular*}{\textwidth}{lrrrrr}
            \toprule[1.5pt]
             & \bf Naturalness & \bf Attentiveness & \bf Consistency & \bf Personality & \bf Knowledgeability\\ \hline
            dataset & $4.20$ $(0.96)$ & $4.22$ $(0.91)$ & $4.36$ $(0.80)$ & $4.55 (0.72)$ & $4.14 (0.97)$ \\ \hline
            random distractors & $4.01$ $(0.80)$ & $3.90$ $(0.93)$ & $4.03$ $(0.73)$ & $3.86$ $(0.65)$ & $3.93$ $(0.72)$ \\
            rule-based distractors & $\mathbf{4.11}$ $(0.67)$ & $\mathbf{4.09}$ $(0.71)$ & $\mathbf{4.05}$ $(0.56)$ & $\mathbf{4.01}$ $(0.69)$ & $\mathbf{4.04}$ $(0.63)$ \\
            \bottomrule[1.5pt]
        \end{tabular*}
    \end{center}
    \caption{\label{t_evaluation} Human evaluation of our baseline model and our dataset. All 5 categories were evaluated on a likert scale with 5 levels. Standard deviation is shown in brackets.}
\end{table*}

\subsubsection{Loss function}
\label{model_lossfunction}
In addition to the language modeling loss, described in section \ref{model_pretraining}, the model was tasked with identifying the correct next utterance in four candidate sequences $\bm{x}^{(1)}, \bm{x}^{(2)}, \bm{x}^{(3)}, \bm{x}^{(4)}$. (The rationale for this will be described below.)
The wrong sequences were built by concatenating the dialog history with three different utterances from our dataset. Then they are fed, together with a label $y$, into the model, given a standard classification loss:
\begin{equation}
    L_{\mathrm{clf}} = -\log{P(y|\bm{x}^{(1)},\bm{x}^{(2)},\bm{x}^{(3)},\bm{x}^{(4)};\Theta)}
\end{equation}
The overall loss to optimize is the sum of both, $L_{\mathrm{lm}}$ and $L_{\mathrm{clf}}$ with the language modeling loss being reduced by half. Combining both of these losses can help to improve generalization and to accelerate convergence as shown by \newcite{radford2018improving}.
In addition, the classification loss can help to refuse at inference time generated sequences  which do not fit well as a good answer. This will be explained further in section \ref{model_decoding}

In our first approach these utterances were randomly chosen from different dialogues about the same movie (hereinafter called \emph{random distractors}). In a second step we used the detailed sentiment labels to create wrong utterances that represent a more challenging task. If the correct utterance contains an entity, then false utterances are selected that also contain that entity and have different sentiments, if possible (hereinafter called \emph{rule-based distractors}).

\subsubsection{Decoding} 
\label{model_decoding}
We used beam search decoding with a beam size of 4 to generate sequences at inference time, when no ground truth labels are available. To normalize over the length of the sequences we used:
\begin{equation}
lp(Y)=\frac{(5+|Y|)^{\alpha}}{(5+1)^{\alpha}}
\end{equation}
which is defined in \cite{wu2016google}. With $|Y|$ as the current sequence length and $\alpha=0.6$ as the length normalization coefficient.

In addition to that, we filtered sequences with multiple identical 3-grams at every beam search step to avoid loops like: '\textit{he performed great and he performed great}' which otherwise is a common occurrence in beam search decoding.  

After all possible sequences were found, we combined the generated score with the logits from the classification layer of our model to choose the final sequence. As the classifier loss has learned to distinguish between a correct and two wrong utterances, this gives an additional source for choosing a final beam.

\section{Evaluation}
\label{main_evaluation}

In section \ref{data_validation} we validated our human/human dataset regarding correct usage of the given profiles. Now we want to evaluate the general dialogue quality for both our dataset and the output of the baseline model. As automated metrics are not very meaningful when used to evaluate the quality of dialogues \cite{liu2016not}, we have performed a human evaluation. The results are shown in table \ref{t_evaluation}. First we explain the used metrics and then evaluate the results regarding our dataset and baseline model.

\subsection{Human Evaluation Metrics}
For the human evaluation we used Amazon Mechanical Turk again. To evaluate our dataset, we presented pairs of dialogue and one profile to crowd workers to rate. For our baseline model, we asked crowd-workers to chat about a given movie, but did not mention that their chat partner is a bot. We asked the Turker to rate some statements according to their agreement on a Likert scale with five levels from \textit{strongly disagree} to \textit{strongly agree}. The following statements were used:

\begin{itemize}
\item Naturalness: The conversation felt natural.
\item Attentiveness: It felt like your chat partner was attentive to the conversation.
\item Consistency: The conversation was overall consistent.
\item Personality: The conversation fits well to the intended character described above.
\item Knowledgeability: Your partner replied correct to the asked questions.
\end{itemize}

Crowd-sourced evaluation may be of low quality, if the crowd-worker are not carefully selected and controlled. Therefore we only accepted crowd-worker with a minimum acceptance-rate of $95\%$ and implemented two fake questions to detect persons that answered randomly. We asked for the subject of the conversation (correct answer is always \textit{movies}), as well as the name of the movie they talked about. If one of the questions was answered incorrectly, we rejected that answer. We evaluated 360 dialogues with 95 different crowd-worker across the three tasks.

\subsection{Dataset}
The results for our dataset, shown in Table~\ref{t_evaluation}, are all above $4$ (between \textit{agree} and \textit{strongly agree}), which means that the collected data are judged as natural and consistent dialogues. The high result of $4.55$ for personality is consistent with our validation and confirms adherence with the profiles. This and the results from our validation in section \ref{data_validation} confirm a senseful dataset with correct labels and natural conversations. 

However, the value regarding the knowledgeability is slightly lower as the others. One downside of the movie- and entity restrictions we had while collecting our data is that sometimes the crowd-worker did not know enough about the subjects they were chatting about. If that were true and one asked a random question, their partner was not able to answer this. In general, most of the questions were answered properly though and our model was able to learn this behaviour quite well.

\subsection{Baseline Model}
We evaluated two variants of our baseline model, one trained with randomly sampled distractors, one with rule-based (sentiment-/entity-sampled) ones (see Section~\ref{model_lossfunction} above).
The results are shown in table \ref{t_evaluation}. We also show automated metrics for our model in table \ref{t_ppl}. The rule-based distractors represent a more difficult classification task at training time and outperformed the random distractor approach in the human evaluation. While both models are nearly equal in naturalness and consistency, rule-based distractors lead to significantly better results in personality and knowledgeability.
However, while evaluating both models by our own, we sometimes noticed inconsistencies regarding the opinions. One reason could be that at pre-training the model has learned to condition only on language. As it is much more likely that these utterances were semantically wrong instead of just expressing the wrong sentiment, the model can not learn to distinguish between the different attitudes properly.

With automated metrics, the approach with random distractors has the better perplexity. That contradicts with the human evaluation, but confirms that automatic metrics do not always correlate with human perception. The hits@$n$ metric though, lines well with the human evaluation. To be comparable, at test time we generated the utterances for both models randomly. The improvement for the rule-based distractors at training time shows, that our additional labels are meaningful and can help to improve the classification task.

The overall results show that it is possible to train an end-to-end chatbot that can not only generate natural answers but also reasonable content, while being consistent to a personality (expressed through opinions to facts) and some external knowledge. 

\begin{table}[t]
\ra{1.2}
\begin{center}
    \begin{tabular}{lrrr}
        \toprule[1.5pt]
         \bf model & \bf perplexity & \bf hits@1 & \bf hits@3 \\ \hline
         random distractors & $\mathbf{12.38}$ & $74.22$ & $92.41$ \\
         rule-based distractors & $13.50$ & $\mathbf{79.33}$ & $\mathbf{95.51}$ \\
         \bottomrule[1.5pt]
    \end{tabular}
\end{center}
\caption{\label{t_ppl}Automated metric results on our test set. hits@$n$ means $n$ best response selection from 19 additional wrong utterances.}
\end{table}

\begin{figure}[!h]
\begin{center}
\includegraphics[scale=0.6]{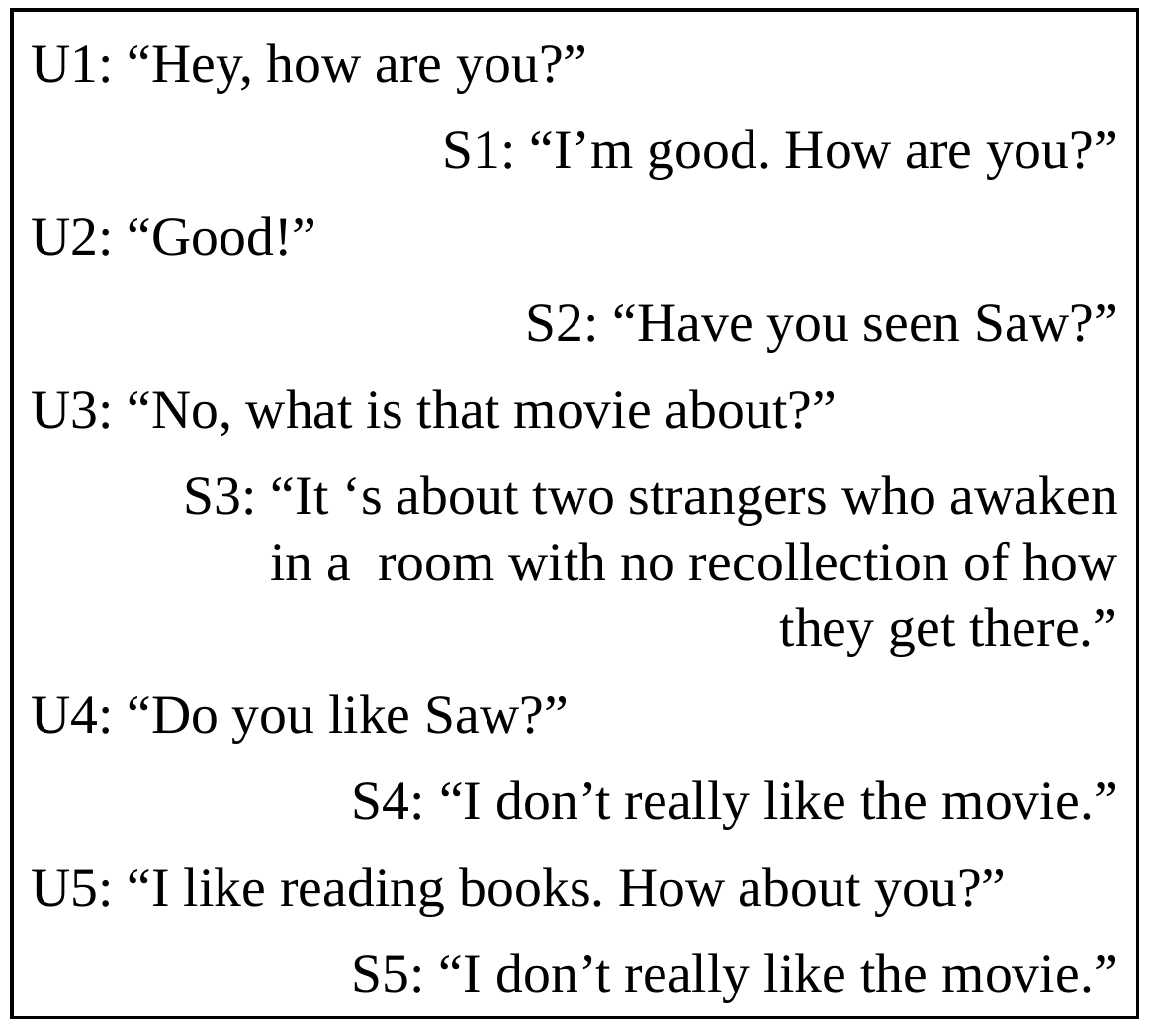} 
\caption{A dialogue between a human and a model trained on our dataset. The conversation about the movie Saw seems quite natural (U1 to S4), but the model is not able to handle different topics (U5, S5).}
\label{fig.model_dials1}
\end{center}
\end{figure}

\begin{figure}[!h]
\begin{center}
\includegraphics[scale=0.6]{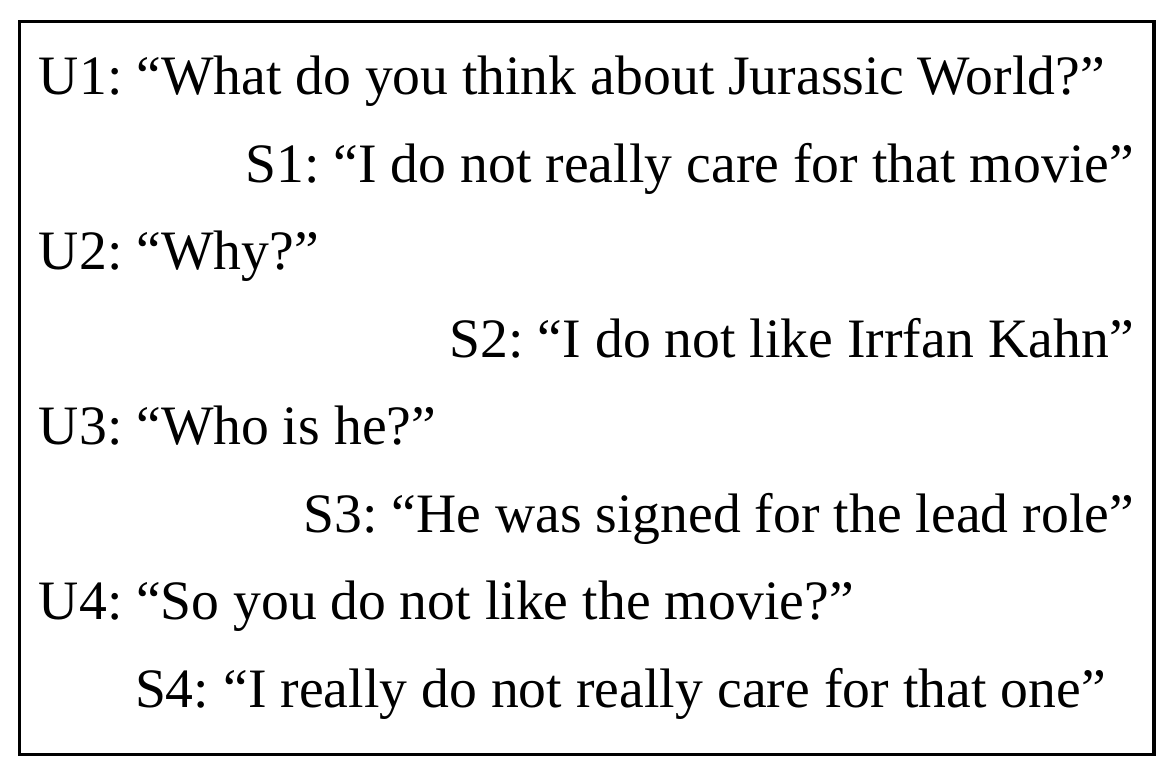} 
\caption{A dialogue with the system trained on our dataset, where the system represents a consistent opinion (S1 to S4). However, the grammar of S4 is wrong.}
\label{fig.model_dials2}
\end{center}
\end{figure}

\section{Conclusion and Future Work}
\label{main_conclusion}
We have presented a new labeled dataset of dialogues, where each dialogue has additional information, namely facts and opinions. This opens a new way to overcome the general problem of inconsistency in end-to-end trained chit-chat models, as we showed with our first baseline model. To be overall consistent, it is important to also be consistently opinionated. With our differentiation of knowledge and opinions, both can be explicitly trained. The baseline model was able to make use of external knowledge in a non-goal driven dialogue, while also representing an opinion and still be natural.

For the future, we are going to explore new model architectures that can handle the additional information in a way different from just concatenating everything as one input sequence. Furthermore, we want to remove the delexicalisation tokens and augment the model with a larger knowledge base, instead of it being restricted to a specific movie. Since our dataset is set in the domain of movies, a model trained on that model is not able to talk about anything outside that domain. It would be interesting to explore if and how it is possible to transfer the property of being opinionated to other, more general dialogue datasets.

\section{Bibliographical References}
\label{main:ref}

\bibliographystyle{lrec}
\bibliography{deeplrec20}


\end{document}